\pdfoutput=1

\documentclass[11pt]{article}

\usepackage{emnlp2021}

\usepackage{times}
\usepackage{latexsym}

\usepackage[T1]{fontenc}

\usepackage[utf8]{inputenc}

\usepackage{microtype}

\usepackage{natbib}
\usepackage{graphicx}
\usepackage{multirow}
\usepackage{amsmath}
\usepackage{booktabs}
\usepackage{subfigure}
\usepackage{epstopdf}
\usepackage{comment}
\usepackage{amssymb}
\usepackage{url}
\usepackage{xcolor}

\usepackage{times}
\usepackage{url}
\usepackage{latexsym}
\usepackage{graphicx}
\usepackage{algorithm}
\usepackage{algorithmic}

\renewcommand{\algorithmiccomment}[1]{\bgroup\hfill//~#1\egroup}

\usepackage[export]{adjustbox}
\usepackage{caption}
\usepackage{booktabs}
\usepackage{multirow}

\title{A Novel Global Feature-Oriented  Relational Triple Extraction Model based on Table Filling}
 \author{Feiliang Ren$^{\dagger,\ast}$, Longhui Zhang$^{\dagger}$,   Shujuan Yin, Xiaofeng Zhao,  Shilei Liu \\  {\bf Bochao Li}, {\bf Yaduo Liu} \\
         School of Computer Science and Engineering \\
         Key Laboratory of Medical Image Computing of Ministry of Education\\ Northeastern University, Shenyang, 110169, China \\ \texttt{renfeiliang@cse.neu.edu.cn}}

\begin{document}
\maketitle
\footnote{\noindent$^\dagger$Both authors contribute equally to this work and share co-first authorship.}
\footnote{$^\ast$Corresponding author.}
\begin{abstract}
Table filling based relational triple extraction methods are attracting growing research interests due to their  promising performance and their abilities on extracting triples from complex sentences. However, this kind of methods are far from their full potential because most of them only focus on using local features  but ignore the global associations of relations and of token pairs, which increases the possibility of overlooking  some important information during triple extraction. To overcome this deficiency, we propose a global feature-oriented triple extraction model that makes full use of the mentioned two kinds of global associations. Specifically, we first generate a table feature for each relation. Then  two kinds of global associations are mined from the generated table features. Next, the mined global associations are integrated into the table feature of each relation. This ``\emph{generate-mine-integrate}'' process is performed multiple times so that the table feature of each relation is refined step by step. Finally, each relation's table is filled based on its refined table feature, and all triples linked to this relation are extracted based on its filled table. We evaluate the proposed model on three benchmark datasets. Experimental results show  our model is effective and it achieves state-of-the-art results on all of these datasets. The source code of our work is available at: https://github.com/neukg/GRTE.
\end{abstract}

\section{Introduction}
Relational triple extraction (RTE)  aims to extract triples  from unstructured text (often sentences), and is  a fundamental task in information extraction. These triples have the form of (\emph{subject, relation, object}) , where both \emph{subject} and \emph{object} are entities and they are semantically linked by \emph{relation}. RTE is important for many downstream applications.

Nowadays, the dominant methods for RTE are the  joint extraction methods that extract entities and relations simultaneously in an end-to-end way.  Some latest joint extraction methods~\citep{yu2019joint,yuan2020a,zeng2020copymtl,wei-etal-2020-novel,wang-etal-2020-tplinker,sun2021} have shown their strong extraction abilities on diverse benchmark datasets, especially the abilities of extracting triples from complex sentences that contain overlapping or multiple triples. 


Among these existing joint extraction  methods,  a kind of table filling based methods~\citep{wang-etal-2020-tplinker,zhang-etal-2017-end,miwa-bansal-2016-end,gupta-etal-2016-table} are attracting growing research attention. These methods usually maintain a table for each relation, and each  item in such a table is used to indicate whether  a token pair possess the corresponding relation   or not.  Thus the key of these   methods  is to fill the relation tables accurately, then the triples can be extracted based on the filled tables. 
However,   existing  methods   fill relation tables mainly based on  local features that are extracted from either a single token pair~\cite{wang-etal-2020-tplinker} or the filled history of some limited token pairs ~\cite{zhang-etal-2017-end}, but  ignore following two kinds of valuable global features: the global associations of  token pairs and  of relations. 

These two kinds of global features can reveal the differences and connections among relations and among token pairs. Thus they are  helpful to both    the precision by verifying the extracted triples from multiple  perspectives, and   the recall by deducing  new triples.  For example, given a sentence ``\emph{Edward Thomas and John are from New York City, USA.}'', when looking it from a global view, we can  easily find following two useful facts. First,   the triple (\emph{Edward Thomas, live\_in, New York}) is  helpful for extracting the triple (\emph{John, live\_in, USA}), and vice versa. This is because the properties of their (\emph{subject, object}) pairs are highly similar: (i) the  types of both  subjects  are  same (both are \emph{persons});  (ii)  the  types of both objects are  same too (both are \emph{locations}). Thus these two entity pairs are highly possible to possess the same kind of relation. Second,  the mentioned two triples  
are helpful for deducing a new triple (\emph{New York, located\_in, USA}). This  is because that: (i) \emph{located\_in} requires both its subjects and objects   be \emph{locations}; (ii) \emph{located\_in} is semantically related to \emph{live\_in}; (iii)  \emph{live\_in}  indicates its objects  are \emph{locations}. Thus there is a clear inference path from these two known triples to the new triple. Obviously, these global features are impossible to be contained in local features. 

Inspired by above analyses,  we propose a global feature-oriented table filling based RTE  model that  fill relation tables mainly based on above two kinds of global associations. In our model, we first generate a table feature for each relation. Then all relations' table features are  integrated into a subject-related global feature and an object-related global feature,  based on which two kinds of global associations are mined with a \emph{Transformer}-based method. Next, these two kinds of mined global associations are used to refine  the table features.  These steps are performed multiple times so that the table features are refined gradually.  Finally, each table is filled based on its refined   feature, and all triples are extracted based on the filled tables.

We evaluate the proposed model on three  benchmark datasets: NYT29, NYT24,  and WebNLG. Extensive experiments show that  it consistently outperforms the existing best models   and achieves the state-of-the-art results on all of these datasets.

\section{Related Work}
Early study~\citep{zelenko2003kernel,zhou2005exploring,chan2011exploiting} often takes a kind of pipeline based methods for RTE, which is to recognize all  entities in the input text first and then to predict the relations for all entity pairs.  However, these methods have two fatal shortcomings. First, they ignore the correlations between  entity recognition and relation prediction. Second, they tend to suffer from the error propagation issue. 

To overcome these shortcomings,   researchers begin to explore the  joint extraction methods that extract entities and relations simultaneously.
According to the research lines taken,  we roughly classify  existing joint methods into three  main kinds. 

\textbf{Tagging based methods}. This kind of methods  ~\citep{zheng2017joint,yu2019joint,wei-etal-2020-novel} often first extract the entities by a tagging based method,  then predict relations. In these models, binary tagging sequences are often used to determine the start and end positions of entities, sometimes to  determine the relations between two entities too. 

\textbf{Seq2Seq based methods}. This kinds of methods~\citep{zeng2018Extracting,zeng2019learning,zeng2020copymtl,DBLP:conf/aaai/NayakN20}  often  view  a triple as a token sequence, and convert the extraction task into a  \emph{generation} task that generates a triple in some orders,  such as first generates a relation, then generates   entities, etc. 

\textbf{Table filling based methods}. This kind of methods ~\citep{miwa-bansal-2016-end,gupta-etal-2016-table,zhang-etal-2017-end,wang-etal-2020-tplinker}  would maintain a table for each relation, and the items in  this table  usually  denotes the start and end positions of two entities (or even the types of these entities) that possess this specific relation. Accordingly, the RTE task is converted into the task of filling these tables accurately and effectively. 

\begin{figure}[t]
	\centering
	\includegraphics[width=0.4\textwidth]{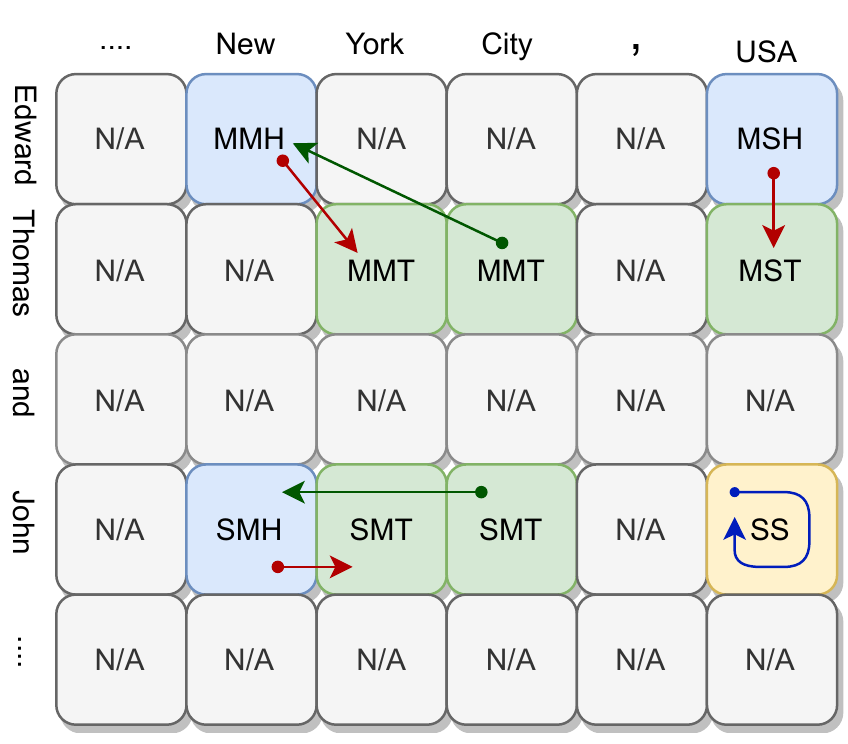}
	\caption{Examples of  table filling and  decoding strategy. Arrows with different colors correspond to different search routes defined in   \emph{Algorithm}~\ref{alg:decoding_1}.  }
	\label{fig:ex}
\end{figure}

Besides,  researchers also explore other kinds of  methods. For example, ~\newcite{bekoulis2018joint}  formulate the RTE task as a multi-head selection problem.  ~\newcite{li2019entity} cast the RTE task as a multi-turn question answering problem. ~\newcite{fu2019graphrel} use  a graph convolutional networks based  method and  ~\newcite{eberts2019span} use a span extraction based method.  ~\newcite{sun2021} propose a multitask learning based RTE model. 

\begin{figure*}[t]
	\centering
	\includegraphics[width=1\textwidth]{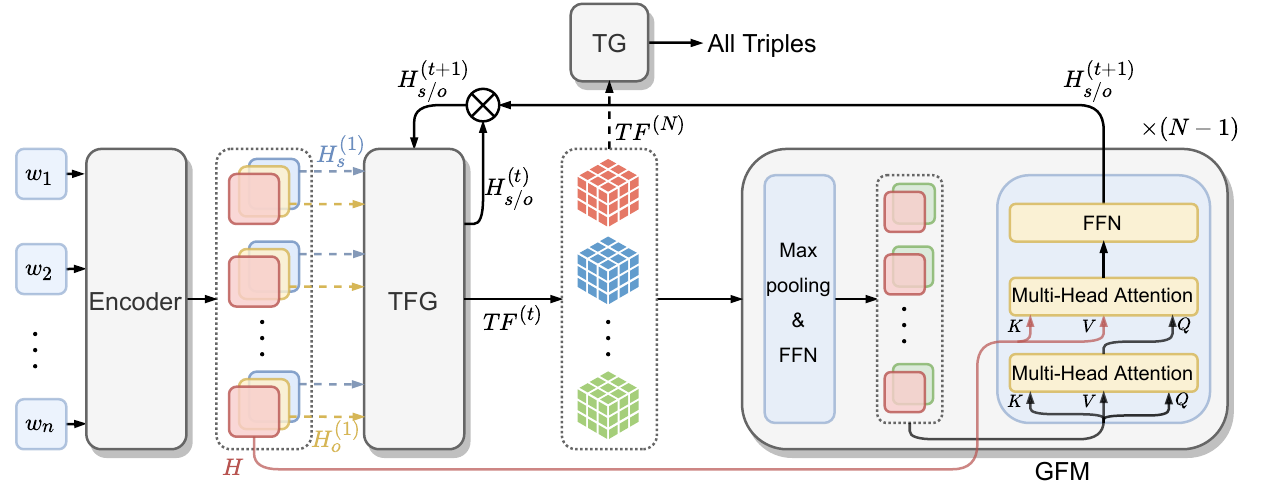}
	\caption{Model Architecture. The dotted arrows to \emph{TFG} means that  $H_s^{(1)}$ and $H_o^{(1)}$  will be inputted to \emph{TFG} only at the first iteration. The dotted arrow to \emph{TG} means that  $TF^{(N)}$ will be inputted  into \emph{TG} only at the last iteration.} 
	\label{fig:IRTE}
\end{figure*}
\section{Methodology}

\subsection{Table Filling Strategy} \label{Tab_Fulling}
Given a sentence $S=w_1w_2\dots w_n$, we will maintain a table $table_{r}$ (the size is ${n \times n}$) for each relation $r$ ($r \in R$, and $R$ is  the relation set). The core of our model is to assign a proper label for each table item  (corresponding to a token pair). Here we define the label set as \emph{L} = \{"N/A", "MMH", "MMT", "MSH", "MST", "SMH", "SMT", "SS"\}. 

For a token pair indexed by the \emph{i}-th row and the \emph{j}-th column, we denote it as ($w_i$, $w_j$) and denote its label as $l$. If $l \in  $  \{"MMH", "MMT", "MSH", "MST", "SMH", "SMT"\}, it means ($w_i$, $w_j$) is correlated with a (\emph{subject, object}) pair. In such case, the first character in the label refers to   the \emph{subject} is either a multi-token entity ("M") or a single-token entity ("S"), the second character in the label refers to  the \emph{object} is either a multi-token entity ("M") or a single-token entity ("S"), and the third character in the label refers to either both $w_i$ and $w_j$ are the head token of the \emph{subject} and \emph{object} ("H") or both are  the tail token of   the \emph{subject} and \emph{object} ("T").  For example, $l$ = "MMH" means  $w_i$ is the head token of a multi-token subject and  $w_j$ is the head token  of  a multi-token object. As for other cases,  $l$ = "SS" means   ($w_i$, $w_j$) is an entity pair;  $l$ = "N/A" means   $w_i$ and  $w_j$  are none of above cases.

Figure~\ref{fig:ex} demonstrates partial filled results for the \emph{live\_in} relation given  the  sentence "\emph{Edward Thomas and John are from New York City, USA.}", where there are  (\emph{subject, object}) pairs of (\emph{Edward Thomas}, \emph{New York City}), (\emph{Edward Thomas}, \emph{New York}), (\emph{Edward Thomas}, \emph{USA}), (\emph{John}, \emph{New York City}), (\emph{John}, \emph{New York}) and (\emph{John}, \emph{USA}). 

An main merit of our  filling strategy is that  each of its label can not only reveal  the location information of a token in a subject or an object, but also  can reveal  whether a subject (or an object) is a single token entity or multi token entity. Thus, the total number of items to be filled under our filling strategy is generally small since the information carried by each label increases.   For example, given a sentence $S=w_1w_2 \dots w_n$ and a relation set $R$, the number of items to be filled under our filling strategy is $n^2|R|$, while this number is $(2|R|+1)\frac{n^{2}+n}{2}$ under the filling strategy used in  \emph{TPLinker}~\citep{wang-etal-2020-tplinker} (this number is copied from the  original paper of \emph{TPLinker} directly). One can easily deduce that  $(2|R|+1)\frac{n^{2}+n}{2} > n^2|R|$.


\subsection{Model Details}

The architecture of our model is shown in Figure~\ref{fig:IRTE}. It consists of four main modules: an \emph{Encoder} module, a \emph{Table  Feature Generation} (\emph{TFG}) module, a \emph{Global  Feature Mining} (\emph{GFM}) module, and a \emph{Triple Generation} (\emph{TG}) module. 
\emph{TFG} and \emph{GFM}  are performed multiple time with an iterative way so that the table features are refined  step by step.   Finally,  $TG$   fills each table based on its corresponding refined table feature  and generates all triples based on these filled tables.






\noindent \textbf{Encoder Module} Here a pre-trained \emph{BERT-Base} (Cased)  model~\citep{devlin2018bert} is used as \emph{Encoder}. Given a sentence, this module  firstly encodes it into a token representation sequence (denoted   as   \emph{H}$\in \mathbb{R}^{n \times d_h}$). 

Then  \emph{H} is fed into two separated \emph{Feed-Forward Networks} (\emph{FFN}) to generate the initial \emph{subjects} feature and \emph{objects} feature (denoted as   $H_s^{(1)}$ and $H_o^{(1)}$ respectively), as   written with Eq.~\eqref{eq:Encoder}. 
\begin{equation}
\begin{aligned}
&H_s^{(1)}=W_1H+b_1\\
&H_o^{(1)}=W_2H+b_2  
\label{eq:Encoder}
\end{aligned}
\end{equation}
where $W_{1/2} \in \mathbb{R}^{d_h \times d_h}$ are trainable  weights and $b_{1/2} \in \mathbb{R}^{d_h}$ are trainable biases. 

\noindent \textbf{TFG Module} We denote the  \emph{subjects} and \emph{objects} features at the $t$-$th$ iteration as $H_s^{(t)}$ and $H_o^{(t)}$ respectively.   Then taking them  as input,  this module generates   a table feature for each relation.  

Here the  table  feature for the relation \emph{r} at the $t$-$th$ iteration is denoted as $TF_{r}^{(t)}$, and it has the same size with $table_r$. Each   item in  $TF_{r}^{(t)}$  represents the label feature for a token pair. Specifically, for a pair $(w_i, w_j)$, we denoted its label feature as $TF_{r}^{(t)}(i,j)$, which  is computed with  Eq.\eqref{eq:TFE}. 
\begin{equation}
TF_{r}^{(t)}(i,j) = W_{r} \operatorname{ReLU}(H_{s,i}^{(t)} \circ H_{o,j}^{(t)}) + b_{r}
\label{eq:TFE}
\end{equation}
where $\circ$ denotes the \emph{Hadamard Product} operation, $\operatorname{ReLU}$ is the activation function, $H_{s,i}^{(t)}$ and $H_{o,j}^{(t)}$ are the feature representations of tokens $w_i$ and $w_j$ at the $t$-$th$ iteration respectively. 

\noindent \textbf{GFM Module} This module mines the expected two kinds of global features, based on which  new \emph{subjects} and \emph{objects}  features are generated. Then these two new generated features will be fed back to   \emph{TFG}  for next  iteration.  Specifically, this module consists of  following  three steps. 


\textbf{Step 1}, to combine table features. Supposing current iteration is \emph{t}, we first concatenate the table features of all relations together to generate an unified table feature (denoted as $TF^{(t)}$). And this unified table feature will contain the information of both token pairs and relations. Then we use  a \emph{max pooling} operation and an \emph{FFN} model on  $TF^{(t)}$ to generate a subject-related table feature ($TF_{s}^{(t)}$) and  an object-related table feature ($TF_{o}^{(t)}$)  respectively, as shown in Eq.\eqref{eq:max_pool}. 
\begin{equation}
\begin{aligned}
&TF_{s}^{(t)}=W_{s} \underset{s}{\operatorname{maxpool}}{\; (TF^{(t)})}+b_{s}\\
&TF_{o}^{(t)}=W_{o} \underset{o}{\operatorname{maxpool}}{\; (TF^{(t)})}+b_{o} 
\label{eq:max_pool}
\end{aligned}
\end{equation}
where $W_{s/o} \in \mathbb{R}^{(|L|\times |R|) \times d_h}$ are trainable weights, and $b_{s/o} \in \mathbb{R}^{d_h}$ are trainable biases.

Here the \emph{max pooling}  is used to highlight the important features that are helpful for the subject  and object extractions respectively from a global perspective.

\textbf{Step 2}, to mine the expected two kinds of global features. Here we mainly use  a \emph{Transformer}-based model~\cite{DBLP:conf/nips/VaswaniSPUJGKP17} to mine the global associations  of relations and of  token pairs.

First, we use a \emph{Multi-Head Self-Attention} method on $TF_{s/o}^{(t)}$  to mine  the global  associations of relations. The self-attention mechanism can reveal the importance of an item from the perspective of other items, thus it is very suitable to mine the expected relation associations. 

Then we mine the global associations of  token pairs with a  \emph{Multi-Head Attention} method. The  sentence representation \emph{H} is also taken as part of input here. We think \emph{H}  may contain some global semantic information of a token to some extent  since the input sentence is encoded as a whole, thus it is helpful for mining the global associations of token pairs from a whole sentence perspective. 

Next, we generate  new  \emph{subjects} and \emph{objects} features with  an \emph{FFN} model. 

In summary, the whole global association mining  process can be written with following Eq.\eqref{eq:transformer}.  
\begin{equation}
\begin{aligned}
& \hat{TF}_{s/o}^{(t)} = \operatorname{MultiHeadSelfAtt}(TF_{s/o}^{(t)}) \\
& \hat{H}_{(s/o)}^{(t+1)}=\operatorname{MultiHeadAtt}(\hat{TF}_{s/o}^{(t)},H,H)\\
&
H_{(s/o)}^{(t+1)}= \operatorname{ReLU}(\hat{H}_{(s/o)}^{(t+1)}W+b)
\label{eq:transformer}
\end{aligned}
\end{equation}

 %


\textbf{Step 3}, to further tune the  \emph{subjects} and \emph{objects} features generated in previous step. 

One can notice that if we flat the iterative modules of \emph{TFG} and \emph{GFM}, our model would equal to a very deep network, thus it is  possible to suffer from the \emph{vanishing  gradient} issue. To avoid this,  we use a \emph{residual network} to generate the final \emph{subjects} and \emph{objects} features, as written in Eq.~\eqref{eq:res}.  
\begin{equation}
H_{(s/o)}^{(t+1)} = \operatorname{LayerNorm}{(H_{(s/o)}^{(t)} + H_{(s/o)}^{(t+1)})} 
\label{eq:res}
\end{equation}

Finally, these  \emph{subjects} and \emph{objects} features are fed back to the  \emph{TFG} module for next iteration. Note that the parameters of \emph{TFG} and \emph{GFM} are \emph{shared} cross different iterations.

\noindent \textbf{TG Module} Taking the table  features at the last iteration ($TF^{(N)}$) as input, this module outputs all the triples. Specifically,  for each relation, its table is  firstly  filled with the method shown in Eq.~\eqref{eq:TFG}. 
\begin{equation}
\begin{aligned}
& \hat{table}_{r}(i,j)=\operatorname{softmax}\;  (TF_{r}^{(N)}(i,j)) \\
&{table}_{r}(i,j)=\underset{l \in  L}{\operatorname{argmax}} \; (\hat{table}_{r}(i,j)[l]) 
\label{eq:TFG}
\end{aligned}
\end{equation}
where  $\hat{table}_{r}(i,j) \in \mathbb{R}^{|L|}$, and  $table_{r}(i,j)$ is the labeled result for the token pair $(w_i,w_j)$ in the table of relation $r$.


Then, \emph{TG} decodes  the filled tables and deduces all  triples   with  Algorithm~\ref{alg:decoding_1}. The  main idea of  our algorithm is  to generate an entity pair set for each relation according to its filled table. And each entity pair in this set would correspond to a minimal continuous token span in the filled table. Then each entity pair would form a triple with the relation that corresponds to the considered table. Specifically, in our decoding algorithm,  we design three paralleled search routes to extract entity pairs of each relation. The first one (\emph{forward search}, red arrows in Figure~\ref{fig:ex})  generates entity pairs in an order of from head tokens  to tail tokens. The second one (\emph{reverse search}, green arrows in Figure~\ref{fig:ex})  generates entity pairs in an order of  from tail tokens to head tokens, which is designed mainly to handle the nested entities.  And the third one (blue arrows in Figure~\ref{fig:ex}) generates entity pairs that are single-token  pairs.  

\begin{algorithm}[t]
	\footnotesize
	\algsetup{
		linenodelimiter = {  }
	}
	\caption{Table Decoding Strategy}
	\label{alg:decoding_1}
	
	\begin{algorithmic}[1]
		\REQUIRE The relation set $R$, the sentence $S$ = \{$w_1$, $w_2$, ..., $w_n$\}, and all $table_{r} \in \mathbb{R}^{n \times n}$ for each relation  $r \in R$.\\
		\ENSURE The predicted triplet set, $RT$.
		\STATE  Define two temporary triple sets \emph{H} and \emph{T}, and initialize $H,T,RT \leftarrow \emptyset,\emptyset,\emptyset$.
		
		
		\FOR{each $r \in R$}
		
		\STATE 
		Define three temporary sets $WP_{r}^{H}$, $WP_{r}^{T}$, and $WP_{r}^{S}$, which consist of token pairs whose ending tags in $table_{r}$ are "H", "T" and "S" respectively. 
		\FOR[forward search]{each $(w_i,w_j) \in WP_{r}^{H}$} 
		\STATE 
		1) Find a token pair $(w_k, w_m)$ from $WP_{r}^{T}$ that satisfies: $i \le k$, $j \le m$, $table_{r}[(w_i,w_j)]$ and $table_{r}[(w_k,w_m)]$ match, $(w_i, w_j)$ and $(w_k, w_m)$ are  closest in the table, and the number of words contained in subject $w_{i \dots k}$ and object $w_{j \dots m}$ are consistent with the corresponding tags. 
		
		\STATE 
		2)		Add $(w_{i \dots k},r,w_{j \dots m})$ to \emph{H}.
		\ENDFOR		
		
		\FOR[reverse search]{each $(w_k,w_m) \in WP_{r}^{T}$}
		\STATE
		1) Find a token pair $(w_i,w_j)$ from $WP_{r}^{H}$ with a similar process as forward search.
		
		\STATE 
		2) Add $(w_{i \dots k},r,w_{j \dots m})$ to \emph{T}.
		\ENDFOR		
		
		\FOR{each $(w_i,w_j) \in WP_{r}^{S}$}
		\STATE Add $(w_i,r,w_j)$ to $RT$
		\ENDFOR
		
		\ENDFOR
		
		\STATE
		$RT \leftarrow RT \cup H \cup T$
		\RETURN $RT$
	\end{algorithmic}
\end{algorithm}

Here we take the sentence shown in Figure~\ref{fig:ex} as a concrete sample to further explain our decoding algorithm. For example, in the demonstrated  table, the token pair (\emph{Edward}, \emph{New}) has an "MMH" label, so the algorithm has to search forward to concatenate adjacent token pairs until a token pair that has a label "MMT" is found, so that to form the complete (\emph{subject}, \emph{object}) pair. And the \emph{forward search} would be stopped when it meets the token pair (\emph{Thomas}, \emph{York}) that has the label "MMT". However, the formed entity pair (\emph{Edward Thomas}, \emph{New York}) is a wrong entity pair in the demonstrated example since the expected pair is (\emph{Edward Thomas}, \emph{New York City}). Such kind of errors are caused by the nested entities in the input sentence, like the ``\emph{New York}'' and ``\emph{New York City}''. These nested entities will make the \emph{forward search} stops too early.  In such case, the designed  \emph{reverse search} will play an important  supplementary role.  In the discussed example, the \emph{reverse search} will first find the token pair (\emph{Thomas}, \emph{City}) that has an "MMT" label and has to further find a token pair that has an "MMH" label. Thus it will precisely find the expected entity pair (\emph{Edward Thomas}, \emph{New York City}). 

Of course, if there are few nested entities in a dataset, the \emph{reverse search}  can be removed, which would be better for the running time. Here we leave  it to make our model have a better generalization ability so that can be used in diverse datasets.



\subsection{Loss Function}
We define the model loss as follows. 
\begin{equation}
\begin{aligned}
\mathcal{L}&=\sum_{i=1}^{n} \sum_{j=1}^{n} \sum_{r=1}^{|R|} -\log p\left(y_{r,(i,j)}=table_{r}(i,j)\right) \\
&=\sum_{i=1}^{n} \sum_{j=1}^{n} \sum_{r=1}^{|R|} -\log \hat{table}_{r}(i,j)[y_{r,(i,j)}]
\end{aligned}
\end{equation}
where $y_{r,(i,j)} \in [1,|L|]$ is the index of the ground truth label of   $(w_i,w_j)$ for the relaion $r$.





\section{Experiments}
\renewcommand{\arraystretch}{1} 
\begin{table}[t] 
	\centering
	\resizebox{0.99\columnwidth}{!}{ 
		\begin{tabular}{lllllll} 
			\toprule
			\multirow{2}{*}{Category}& 
			\multicolumn{2}{c}{NYT29}&\multicolumn{2}{c}{NYT24}&\multicolumn{2}{c}{WebNLG}\\
			\cmidrule(lr){2-3} \cmidrule(lr){4-5} \cmidrule(lr){6-7}&Train&Test&Train&Test&Train&Test\\
			\midrule 
			Normal  &53444& 2963& 37013& 3266& 1596& 246              \\ 
			EPO    &8379& 898 & 9782& 978 & 227& 26              \\ 
			SEO    & 9862&1043& 14735&  1297& 3406& 457              \\ 
			
			\midrule
			ALL&63306 &4006&56195&5000&5019&703\\
			Relation &\multicolumn{2}{c}{29}&\multicolumn{2}{c}{24}&\multicolumn{2}{c}{216 / 171$^*$}\\
			\bottomrule 
		\end{tabular}
	}
	\caption{Statistics of datasets. \emph{EPO} and \emph{SEO} refer to  \emph{entity pair overlapping} and \emph{single entity overlapping} respectively~\citep{zeng2018Extracting}. Note that a sentence can 	belong to both \emph{EPO}  and \emph{SEO}. And 216 / 171$^*$ means that there are 216 / 171 relations in WebNLG and  WebNLG$^*$ respectively. }
	\label{tab:statistics} 
\end{table}

		

\subsection{Experimental Settings}
\noindent \textbf{Datasets}  We evaluate our model on three benchmark  datasets: NYT29~\citep{takanobu2019a}, NYT24~\cite{zeng2018Extracting} and WebNLG~\citep{gardent2017Creating}. Both NYT24 and WebNLG have two different versions according to following two annotation  standards: 1) annotating the last token of each entity, and 2) annotating the whole entity span. Different work chooses different versions of these datasets. To evaluate our model comprehensively, we use both kinds of datasets. For convenience,  we denote the datasets based on the first annotation standard as NYT24$^*$ and WebNLG$^*$, and the datasets based on the second annotation standard as NYT24 and WebNLG. Some statistics of these datasets are shown in Table~\ref{tab:statistics}.

\noindent \textbf{Evaluation Metrics} The standard micro precision, recall, and \emph{F1} score are used to evaluate the results.

Note that there are two match standards for the RTE task:  one is \emph{Partial Match} that an extracted triplet is regarded as correct if the predicted relation and the head tokens of both subject entity and object entity are  correct; and the other is \emph{Exact Match} that a  triple would be  considered  correct  only when its  entities and relation are completely matched with a correct triple. To fairly compare with existing models, we follow previous work~\cite{wang-etal-2020-tplinker,wei-etal-2020-novel,sun2021} and use \emph{Partial Match} on NYT24$^*$ and WebNLG$^*$, and use \emph{Exact Match} on NYT24, NYT29, and WebNLG. 

In fact, since only one token of each entity in NYT24$^*$ and WebNLG$^*$ is annotated, the results of \emph{Partial Match} and \emph{Exact Match} on these two datasets are actually the same. 


\noindent \textbf{Baselines} We compare our model with  following strong state-of-the-art RTE models:
\emph{CopyRE}~\cite{zeng2018Extracting}, \emph{GraphRel}~\cite{fu2019graphrel}, 	 \emph{CopyMTL}~\cite{zeng2020copymtl},	  \emph{OrderCopyRE}~\cite{zeng2019learning}, \emph{ETL-Span}~\cite{yu2019joint}, \emph{WDec}~\cite{DBLP:conf/aaai/NayakN20}, \emph{RSAN}~\cite{yuan2020a}, \emph{RIN}~\cite{DBLP:conf/emnlp/SunZMML20}, \emph{CasRel}~\cite{wei-etal-2020-novel}, \emph{TPLinker}~\cite{wang-etal-2020-tplinker}, \emph{SPN}~\cite{DBLP:journals/corr/abs-2011-01675}, and \emph{PMEI}~\cite{sun2021}.  

Most of the experimental results  of these baselines  are copied from their original papers directly. Some baselines did not report their results on some of the used  datasets. In  such case, we report the best results we obtained with the provided source code (if the source codes is available). For simplicity, we denote our model as \emph{GRTE}, the abbreviation of \emph{G}lobal feature oriented \emph{RTE} model.

\noindent \textbf{Implementation Details} Adam~\cite{KingmaB14} is used to optimize \emph{GRTE}. The learning rate, epoch and batch size are set to 3$\times$$10^{-5}$, $50$, $6$ respectively. The iteration numbers (the hyperparameter $N$) on NYT29, NYT24$^*$, NYT24, WebNLG$^*$ and WebNLG  are set to 3, 2, 3, 2, and 4  respectively. 
Following previous work~\cite{wei-etal-2020-novel,sun2021,wang-etal-2020-tplinker}, we also implement a \emph{BiLSTM}-encoder version of \emph{GRTE} where   300-dimensional GloVe embeddings~\cite{pennington2014Glove} and 2-layer stacked BiLSTM are used. In this version, the hidden dimension of these 2 layers are set as 300 and 600 respectively.  All the   hyperparameters reported in this work are determined based on the results on the development sets. Other parameters are randomly initialized. Following \emph{CasRel} and \emph{TPLinker}, the max length of input sentences is set to 100.

\begin{table*}[htbp]
	\small
	\begin{center}

		\setlength{\tabcolsep}{1.3 mm}
		{\begin{tabular}{lcccccccccccccccc}
				\toprule
				\multirow{2}{*}{Model} & \multicolumn{3}{c}{NYT29} &\multicolumn{3}{c}{NYT24$^\star$} & \multicolumn{3}{c}{NYT24}& \multicolumn{3}{c}{WebNLG$^\star$} & \multicolumn{3}{c}{WebNLG}\\
				& Prec. & Rec. & F1 & Prec. & Rec. & F1 & Prec. & Rec. & F1 & Prec. & Rec. & F1 & Prec. & Rec. & F1 \\
				\midrule
				CopyRE & -- & -- & -- & 61.0 & 56.6 & 58.7 & -- & -- & -- & 37.7 & 36.4 & 37.1 & -- & -- & --  \\
				GraphRel & -- & -- & -- & 63.9 & 60.0 & 61.9 & -- & -- & -- & 44.7 & 41.1 & 42.9 & -- & -- & -- \\
				OrderCopyRE & -- & -- & -- & 77.9 & 67.2 & 72.1 & -- & -- & -- & 63.3 & 59.9 & 61.6 & -- & -- & --  \\
				ETL-Span &74.5$^\star$ &57.9$^\star$ &65.2$^\star$ & 84.9 & 72.3 & 78.1 & 85.5 & 71.7 & 78.0 & 84.0 & 91.5 & 87.6 & 84.3  & 82.0 & 83.1 \\
				WDec &77.7 &60.8 &68.2 & -- & -- & -- &88.1 &76.1 &81.7 & -- & -- & -- & -- & -- & --  \\
				RSAN&--&--&--&--&--&--&85.7 &83.6 &84.6 &--&--&--&80.5 &83.8 &82.1\\
				
				RIN & -- & -- & -- &87.2 &87.3 &87.3 &83.9 &85.5 &84.7 &87.6 &87.0 &87.3 &77.3 &76.8 &77.0 \\
				
				CasRel$_{LSTM}$ &--&--&-- & 84.2 & 83.0 & 83.6 & -- & -- & -- & 86.9 & 80.6 & 83.7 & -- & -- & --  \\
				PMEI$_{LSTM}$ &--&--&-- &88.7 &86.8 &87.8 &84.5 &84.0 &84.2 &88.7 &87.6 &88.1  &78.8 &77.7 &78.2 \\
				TPLinker$_{LSTM}$ & -- & -- & -- & 83.8 & 83.4 & 83.6 & 86.0 & 82.0 & 84.0 & 90.8 & 90.3 & 90.5 & 91.9  & 81.6 & 86.4 \\
				
				CasRel$_{BERT}$ &77.0$^\star$ &{68.0}$^\star$ &72.2$^\star$ & 89.7 & 89.5 & 89.6 &89.8$^\star$ &88.2$^\star$ &89.0$^\star$ & 93.4 & 90.1 & 91.8 &88.3$^\star$ &84.6$^\star$ &86.4$^\star$  \\
				PMEI$_{BERT}$ & -- & -- & -- &90.5 &89.8 &90.1 &88.4 &88.9 &88.7 &91.0 &92.9 &92.0 &80.8 &82.8 &81.8 \\
				TPLinker$_{BERT}$ &78.0$^*$ & 68.1$^*$ & 72.7$^*$ & 91.3 &92.5 & 91.9 & 91.4 &92.6 & 92.0 & 91.8 & 92.0 &91.9 & 88.9 &84.5 & 86.7  \\				
				SPN$_{BERT}$ &76.0$^*$ &71.0$^*$ &73.4$^*$ &\textbf{93.3} &91.7 &92.5 &92.5 &92.2 &92.3 &93.1 &93.6 &93.4 &85.7$^\star$ & 82.9$^\star$ &84.3$^\star$  \\
				\midrule
				GRTE$_{LSTM}$  &74.3&67.9&71.0 &87.5&86.1&86.8&86.2&87.1&86.6&90.1&91.6&90.8&88.0&86.3&87.1\\
				GRTE$_{BERT}$ &\textbf{80.1}&\textbf{71.0}&\textbf{75.3} &92.9     &\textbf{93.1} &\textbf{93.0}  &\textbf{93.4}   &\textbf{93.5} &\textbf{93.4}   &\textbf{93.7}  &\textbf{94.2} &\textbf{93.9}  &\textbf{92.3}  &\textbf{87.9} &\textbf{90.0}\\	
				\midrule
				
				GRTE$_{w/o\;GFM}$&77.9&68.9&73.1 &90.6&92.5&91.5&91.8&92.6&92.2&92.4&91.1&91.7&88.4&86.7&87.5\\
				GRTE$_{GRU \;GFM}$&78.2&\textbf{71.7}&74.8 &92.5&92.9&92.7&93.4&92.2&92.8&93.4&92.6&93.0&90.1&\textbf{88.0}&89.0\\

				GRTE$_{w/o \;m-h}$&77.8&70.9&74.2 &91.9&92.9&92.4&93.2&92.9&93.0&92.9&92.1&92.5&90.5&87.6&89.0\\
				GRTE$_{w/o \; shared}$ &79.5&71.5&\textbf{75.3}&92.7&93.0&92.8&\textbf{93.6}&92.7&93.1&93.4&94.0&93.7&91.5&87.4&89.4\\
				\bottomrule
		\end{tabular}}
		{
			\caption{Main results. A model with a subscript \emph{LSTM} refers to replace its \emph{BERT} based encoder with the \emph{BiLSTM} based encoder.   $^\star$ means the results are  produced by us with the available source code.  }
			\label{table:main_res}
		} 
	\end{center}
\end{table*}

\subsection{Main Experimental Results}

The main results are in the top two parts of Table~\ref{table:main_res}, which show \emph{GRTE} is very effective. On all datasets, it  achieves almost all  the best results  in term of \emph{F1}  compared with the models that use the same kind of encoder (either the BiLSTM based encoder or the BERT based encoder). The only exception is on NYT24$^*$, where the \emph{F1}  of \emph{GRTE}$_{LSTM}$ is about 1\% lower than that of \emph{PMEI}$_{LSTM}$. However, on the same dataset, the \emph{F1} score of \emph{GRTE}$_{BERT}$ is about 2.9\% higher than that of \emph{PMEI}$_{BERT}$. 

The results also show that  \emph{GRTE} achieves much better results on NYT29, NYT24 and WebNLG:  its \emph{F1} scores improve about 1.9\%, 1.1\%, and 3.3\%  over the previous best models on these three datasets respectively.  Contrastively,  its \emph{F1} scores improve about 0.5\% and 0.5\% over the  previous best models on NYT24$^*$ and WebNLG$^*$ respectively. This is mainly because  that \emph{GRTE} could not realize its full potential on NYT24$^*$ and WebNLG$^*$ where only one token of each entity is annotated. For example, under this annotation standard,  except {"N/A", "SSH", and "SST"}, all the other defined labels in \emph{GRTE} are redundant.  But it should be noted that the annotation standard on NYT24$^*$ and WebNLG$^*$ simplifies the RTE task, there would not be such a standard when a model is really deployed. Thus, the annotation  standard on NYT29, NYT24 and WebNLG can better reveal the true performance of a model.  Accordingly, \emph{GRTE}'s better performance on them is more meaningful. 



We can further see that   compared with the previous best models,  \emph{GRTE} achieves more  performance improvement on WebNLG   than on other datasets. For example,  \emph{GRTE}$_{LSTM}$ even outperforms all other compared baselines on WebNLG, including those models that use \emph{BERT}. We think this is mainly because that the numbers of relations in WebNLG are far more than those of in  NYT29 and NYT24 (see Table~\ref{tab:statistics}), which means  there are more  global associations of relations can be mined. 
Generally, the more relations and entities there are in a dataset, the more global correlations there would be among triples. Accordingly, our model could perform more better on such kind of  datasets than other local features based methods. For example,  the number of relations in WebNLG is almost 7 times of those in NYT, and  \emph{GRTE} achieves much more performance improvement over the compared baselines on WebNLG than on NYT. 




\subsection{Detailed Results}
In this section, we conduct detailed experiments to  demonstrate the effectiveness of our model from following two aspects. 

\textbf{First}, we conduct some ablation experiments to  evaluate the  contributions of some main components in \emph{GRTE}. To this end, we  implement following model variants.    

(i) GRTE$_{w/o\;GFM}$, a variant that removes the \emph{GFM} module completely from \emph{GRTE}, which is to evaluate the contribution of \emph{GFM}. Like previous table filling based methods, GRTE$_{w/o\;GFM}$ extracts triples only based on local features. 


(ii) GRTE$_{GRU\;GIF}$, a variant that uses \emph{GRU} (taking $H$ and   $TF_{s/o}^{(t)}$ as  input) instead of  \emph{Transformer}  to generate the results in Eq.~\eqref{eq:transformer}, which is to evaluate the contribution of \emph{Transformer}. 

\begin{table*}[th]
	\small
	\begin{center}
		
		\setlength{\tabcolsep}{0.5 mm}
		{\begin{tabular}{lccccccccccccccccc}
				\toprule
				\multirow{2}{*}{Model} & \multicolumn{8}{c}{NYT24$^\star$}& \multicolumn{9}{c}{WebNLG$^\star$} \\
				& Normal & SEO & EPO & T = 1 & T = 2 & T = 3 & T = 4 & T $\geq$ 5 && Normal & SEO & EPO & T = 1 & T = 2 & T = 3 & T = 4 & T $\geq$ 5 \\
				\midrule
				CasRel$_{BERT}$  & 87.3 & 91.4 & 92 & 88.2 & 90.3 & 91.9 & 94.2 & 83.7 &&89.4 & 92.2 & 94.7 &89.3 &90.8 & 94.2 & 92.4 & 90.9  \\
				TPLinker$_{BERT}$  & {90.1} & {93.4} & {94.0} & {90.0} & {92.8} & {93.1} & {96.1} & {90.0} && 87.9 & {92.5} & {95.3} & 88.0 & 90.1 & {94.6} & {93.3} & {91.6}\\ 
				SPN$_{BERT}$  &90.8 &94.0 &94.1 &\textbf{90.9} &93.4 &94.2 &95.5 &90.6 &&89.5$^*$&94.1$^*$&90.8$^*$&89.5 &91.3 &96.4 &94.7 &93.8\\
				\midrule
				GRTE$_{BERT}$  &\textbf{91.1} &\textbf{94.4} &\textbf{95} &90.8&\textbf{93.7}&\textbf{94.4}&\textbf{96.2}&\textbf{93.4}&&\textbf{90.6}&\textbf{94.5}&\textbf{96}&\textbf{90.6}&\textbf{92.5}&\textbf{96.5}&\textbf{95.5}&\textbf{94.4}\\
				\bottomrule
		\end{tabular}}
		\caption{F1 scores on sentences with different overlapping pattern and different triplet number. Results of \emph{CasRel} are copied from \emph{TPLinker} directly. ``T'' is the number of triples contained in a sentence. $^*$ means the results are produced by us with the provided source codes.
		}
		\label{table:f1_on_split}
	\end{center}
\end{table*}

(iii) GRTE$_{w/o\;m-h}$, a variant that  replaces the \emph{multi-head attention} method in \emph{GFM}  with a \emph{single-head attention} method, which is to evaluate the contribution of the \emph{multi-head attention}.

(iv) GRTE$_{w/o\;shared}$, a variant that uses different  parameters for the modules of \emph{TFG} and \emph{GFM}  at different iterations, which is to evaluate the contribution of the parameter share mechanism.


All these variants use the \emph{BERT}-based encoder. And their  results  are shown in the bottom part of  Table~\ref{table:main_res}, from which we can make following observations.

(1) The performance of GRTE$_{w/o\;GFM}$  drops  greatly compared with \emph{GRTE},  which   confirms the importance of using two kinds of global features for table filling.   We can further  notice that on NYT29, NYT24, and WebNLG, the \emph{F1} scores of GRTE$_{w/o\;GFM}$ increases by 0.4\%, 0.4\%, and 0.8\% respectively over \emph{TPLinker}. Both \emph{TPLinker} and GRTE$_{w/o\;GFM}$ extract triples based on local features, and the main difference between them  is the table filling strategy. So these results prove the effectiveness of our table filling strategy. The \emph{F1} scores of GRTE$_{w/o\;GFM}$  on NYT24$^*$ and WebNLG$^*$ are slightly lower than those of  \emph{TPLinker}, as explained above,  this is because  each entity  in NYT24$^*$ and WebNLG$^*$, only one token is annotated for each entity,  GRTE$_{w/o\;GFM}$ could not realize its full potential.



(2) The performance of GRTE$_{GRU\;GFM}$ drops  compared with \emph{GRTE}, which  indicates   \emph{Transformer}   is more suitable for the global feature mining  than  \emph{GRU}. But even so, we can see that  on all datasets, GRTE$_{GRU\;GFM}$ outperforms almost all   previous best models and  GRTE$_{w/o\;GFM}$ in term of   \emph{F1},  which further indicates the effectiveness of using   global features.  

(3) The results of GRTE$_{w/o\;m-h}$ are lower than those of  \emph{GRTE}, which  shows the \emph{multi-head attention} mechanism plays an important role for global feature mining. In fact, the importance of different features is different,  the \emph{multi-head attention} mechanism performs the  feature mining process from multiple aspects, which is much helpful to highlight the more important ones.

(4) The results of GRTE$_{w/o\;shared}$ are slightly lower than those of \emph{GRTE}, which  shows the  share mechanism is effective. In fact, the mechanism of using distinct parameters  usually works well only when the training samples are sufficient. But this condition is not well satisfied in RTE since the training samples of a dataset are not sufficient enough to train too many parameters.

\begin{figure}[t]
	\centering
	\includegraphics[width=0.45\textwidth]{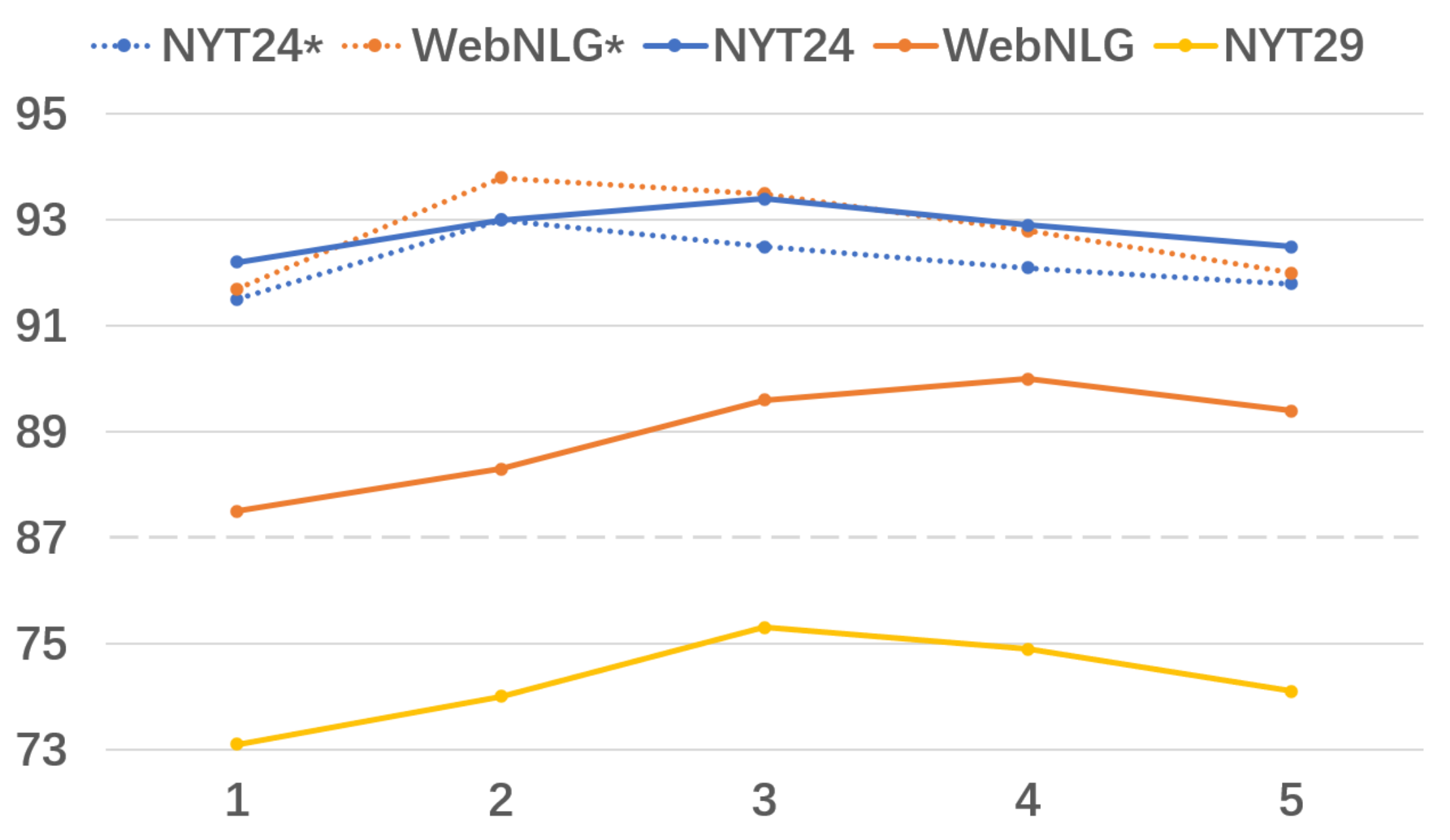}
	\caption{F1 results under different \emph{N}.}
	\label{fig:tf_num}
\end{figure}

\textbf{Second}, we evaluate the influence of the iteration number $N$. The results are shown in Figure~\ref{fig:tf_num}, from which   following observations can be made.

(1) On NYT24$^*$ and WebNLG$^*$, the annotation  standard is relatively simple. So \emph{GRTE} achieves the best results with  two iterations. But on NYT29, NYT24, and WebNLG, more iterations are usually required. For example, \emph{GRTE} achieves the best results when \emph{N} is 3, 3, and 4 respectively on them. 

(2) On all datasets, \emph{GRTE} gets  obvious  performance improvement (even the maximum performance improvement on some datasets) at $N = 2$ where \emph{GFM} begins to  play its role , which  indicates again that using global features can significantly improve the model performance. 
       
(3) \emph{GRTE} usually  achieves the best results within a small number of iterations on all datasets including WebNLG or WebNLG$^*$ where there are lots of relations. In fact,  \emph{GRTE} outperforms all the pervious best models  even when \emph{N} = 2. This  is a very important merit because it indicates that even used on some  datasets where the numbers of relations are very large, the \emph{efficiency}  would not be a burden for \emph{GRTE}, which is much meaningful when \emph{GRTE} is deployed in some real scenarios. 

\begin{table*}[th]
	\small
	\begin{center}
		
		\setlength{\tabcolsep}{1.5mm}{\begin{tabular}{lccccccccc}
				\toprule
				\multirow{2}{*}{Model} 
				& \multicolumn{3}{c}{NYT24$^\star$}
				& \multicolumn{3}{c}{WebNLG$^\star$}\\
				& Params$_{all}$ & Prop$_{encoder}$ & Inference Time & Params$_{all}$ & Prop$_{encoder}$ & Inference Time \\
				\midrule
				CasRel$_{BERT}$ & 107,719,680
				& 99.96\% & 53.9 & 107,984,216 & 99.76\%  & 77.5 \\
				
				TPLinker$_{BERT}$ & 109,602,962 & 98.82\%  & 18.1 / 83.5$^\dagger$ & 110,281,220 & 98.21\% & 26.9 / 120.4$^\dagger$ \\
				SPN$_{BERT}$ &141,428,765 &76.58\%   &26.4 / 107.9$^\dagger$  &150,989,744 &71.73\%  &22.6 / 105.7$^\dagger$ \\
				\midrule
				GRTE$_{BERT}$ &119,387,328&90.72\%&21.3 / 109.6$^\dagger$ &122,098,008&88.70\%&28.7 / 124.1$^\dagger$  \\

				\bottomrule
		\end{tabular}}
		{
			\caption{Computational efficiency. Params$_{all}$ is the number of parameters for the entire model. Prop$_{encoder}$ refers to the proportion of encoder parameters in the total model parameters. Inference Time represents the average time (millisecond) the model takes to process a sample. $^\dagger$ marks the inference time when the batch size is set to 1. }
			\label{table:inference time}
		} 
	\end{center}
\end{table*}


\subsection{Analyses on Different Sentence Types}

Here we  evaluate \emph{GRTE}'s  ability for extracting triples from sentences that contain overlapping triples and multiple triples.  For fair comparison with the previous best models (\emph{CasRel}, \emph{TPLinker}, and \emph{SPN}), we follow their  settings which are: (i) classifying sentences according to the degree of overlapping and the number of triples contained in a sentence, and (ii) conducting experiments on different subsets of NYT24$^*$ and WebNLG$^*$. 

The results are shown in Table~\ref{table:f1_on_split}. We can see  that: (i) \emph{GRTE} achieves the best results on all three kinds of overlapping sentences on both datasets, and (ii) \emph{GRTE} achieves the best results on almost  all kinds of  sentences that contain multiple triples. The only exception is on  NYT24$^*$ where the \emph{F1} score of \emph{GRTE} is slightly lower than that of  \emph{SPN}  when \emph{T} is 1. The main reason is that  there are less associations among token pairs when \emph{T} is 1, which slightly degrades the performance of \emph{GRTE}. 


In fact,  \emph{GRTE} maintains a table for each relation, and the \emph{TG} module extracts triples for each relation  independently. Thus it can well handle  above two kinds of complex sentences by nature.

\subsection{Analyses on Computational Efficiency}
Table~\ref{table:inference time} shows the comparison results of  computational efficiency between \emph{GRTE} and  some previous best models. To be fair,  we follow the settings in  \emph{TPLinker}:   analyze the parameter scale and the inference time  on NYT$^*$ and WebNLG$^*$. All the  results are obtained by running the compared models on a TitanXP, and  the batch size is set to 6 for all  models that can be run in a batch mode. 

The parameter number of \emph{GRTE} is slightly larger than that of \emph{TPLinker}, which is mainly   due to the using of  a  \emph{Transformer}-based model. But when compared with \emph{SPN} that uses the \emph{Transformer} model too, we can see that \emph{GRTE} has a  smaller number of parameters due to its  parameter share mechanism. 

We can also see that \emph{GRTE} achieves a very competitive inference speed. This is mainly because of following three reasons. First, \emph{GRTE}  is a one-stage extraction model and  can process samples in a batch mode (\emph{CasRel}   can only process samples one by one). Second, as analyzed previously, it has an efficient  table filling strategy that needs to fill fewer table items. Third, as analyzed previously, \emph{GRTE} often achieves the best results within a small number of iterations, thus the iteration operations  will not have too much impact on the  inference speed of \emph{GRTE}. 

In fact, as \emph{TPLinker} pointed out  that for all the models that use BERT (or other kinds of pre-trained language models) as their basic encoders, BERT is usually the most time-consuming part and takes up the most of model parameters, so the time cost of other components in a model is not significant. 

Besides, there is another important merit of our model: it  needs less training time than existing state-of-the-art models like \emph{CasRel}, \emph{TPLinker}, and \emph{SPN} etc. As pointed out previously, the epoch of our model on all datasets is 50. But on the same datasets, the epochs of all the mentioned models are 100. From Table~\ref{table:inference time} we can see that all these models have similar inference speed. For each model, the training speed  of each epoch is very close to its inference speed (during training, there would be extra time cost for operations like the back propagation), thus we can easily know that our model needs less time for training since our model has a far less epoch number.

\section{Conclusions}
In this study, we propose a novel  table filling based  RTE model that extracts triples based on two kinds of global features. The main contributions of our work are listed as follows. 
First,  we make use of   the global associations of relations and of token pairs. Experiments show these two kinds of global features  are much helpful for performance. 
Second, our model  works well on extracting triples from  complex sentences  containing overlapping triples or    multiple triples. 
Third, our model is evaluated on three benchmark datasets. Extensive experiments  show that it consistently outperforms all the compared strong baselines and achieves  state-of-the-art results. Besides, our model has a   competitive inference speed and a moderate parameter size.

\section*{Acknowledgments}
This work is supported by the National Natural Science Foundation of China (No.61572120 and No.U1708261), the Fundamental Research Funds for the Central Universities (No.N181602013 and No.N2016006), Shenyang Medical Imaging Processing Engineering Technology Research Center (17-134-8-00),  Ten Thousand Talent Program (No.ZX20200035), and Liaoning Distinguished Professor (No.XLYC1902057).

\bibliography{anthology,custom}
\bibliographystyle{acl_natbib}

\end{document}